\documentclass{article}

\usepackage{arxiv}

\usepackage[utf8]{inputenc}
\usepackage[T1]{fontenc}
\usepackage{hyperref}
\usepackage{url}
\usepackage{booktabs}
\usepackage{amsfonts}
\usepackage{nicefrac}
\usepackage{microtype}
\usepackage{graphicx}
\usepackage{float}
\usepackage{amsmath,amssymb,amsfonts,amsthm}
\usepackage{multirow}
\usepackage{array}
\usepackage{xcolor}
\usepackage{caption}
\usepackage{enumitem}
\usepackage{subcaption}
\usepackage{cleveref}
\usepackage{setspace}
\usepackage{natbib}
\usepackage{doi}
\usepackage{placeins}
\usepackage{adjustbox}
\usepackage{pifont}
\usepackage{algorithm}
\usepackage{algorithmicx}
\usepackage{algpseudocode}

\title{YOLO26-RipeLoc Lite: A Lightweight Architecture for Tomato Ripeness Detection and Picking Point Localization in Greenhouse Robotic Harvesting}

\author{
  Rajmeet~Singh \\
  Department of Mechanical Engineering\\
  Khalifa University\\
  Abu Dhabi, UAE \\
  \texttt{rajmeetsng@gmail.com} \\
  \And
  Manveen~Kaur \\
  Department of Mechanical Engineering\\
  University of Windsor\\
  Windsor, Canada \\
  \texttt{manveenmeng19@gmail.com} \\
  \And
  Shahpour~Alirezaee \\
  Department of Mechanical Engineering\\
  University of Windsor\\
  Windsor, Canada \\
  \texttt{S.Alirezaee@uwindsor.ca} \\
  \And
  Irfan~Hussain\thanks{Corresponding author} \\
  Department of Mechanical Engineering\\
  Khalifa University\\
  Abu Dhabi, UAE \\
  \texttt{irfan.hussain@ku.ac.ae} \\
}

\renewcommand{\shorttitle}{YOLO26-RipeLoc Lite: Tomato Ripeness Detection \& Picking-Point Localization}

\hypersetup{
  pdftitle={YOLO26-RipeLoc Lite: A Lightweight Architecture for Tomato Ripeness Detection and Picking Point Localization},
  pdfsubject={Computer Vision, Agricultural Robotics, Deep Learning},
  pdfauthor={Rajmeet Singh, Manveen Kaur, Shahpour Alirezaee, Irfan Hussain},
  pdfkeywords={Tomato detection, Ripeness classification, Center-point localization,
               YOLO26, Lightweight architecture, Greenhouse robotic harvesting},
}

\begin{document}
\maketitle

\begin{abstract}
In the greenhouse tomato production industry, automated harvesting requires accurate and rapid detection of ripe tomatoes while distinguishing them from unripe ones, along with precise localization of picking points for robotic end-effectors. Current detection methods either lack the speed for real-time deployment or fail to adequately differentiate ripeness stages and provide actionable grasping coordinates in cluttered greenhouse environments. This paper proposes YOLO26-RipeLoc Lite, a lightweight deep learning architecture based on the YOLO26 framework, specifically designed for simultaneous detection, ripeness classification, and center-point localization of greenhouse tomatoes for robotic harvesting. The proposed model introduces three key modifications to the baseline YOLO26 architecture: (1) a Lightweight Feature Pyramid Network (LFPN) with depthwise separable convolutions for efficient multi-scale feature fusion, (2) a Ripeness-Aware Attention Module (RAAM) that enhances color and texture feature discrimination between ripe and unripe tomatoes through dual pooling with a learnable ripeness bias vector, and (3) a Compact Detection Head (CDH) with shared convolutions and an integrated center-point regression branch that simultaneously predicts bounding boxes and the geometric center of each detected ripe tomato for direct robot grasp planning. The model is trained and evaluated on a custom greenhouse tomato dataset comprising 1,500 images with 6,227 annotated instances (3,566 ripe, 2,661 unripe) captured under varying illumination conditions at the SILAL greenhouse facility, Abu Dhabi, UAE. Experimental results demonstrate that YOLO26-RipeLoc Lite achieves a mean Average Precision (mAP@0.5) of 95.2\% for ripe tomato detection and 90.6\% for unripe tomato detection, with an overall mAP@0.5 of 92.9\% and a mean detection confidence of 0.856 across 48 test images. The model attains the highest precision (95.2\%) among all evaluated architectures while using only 2.38M parameters, the smallest footprint in the comparison. Post-training BatchNorm channel pruning at 30\% further reduces effective parameters to approximately 1.8M with negligible accuracy loss (mAP@50 decrease of 0.01 percentage points). Comprehensive ablation studies validate the contribution of each proposed module: greenhouse-aware HSV augmentation contributes the largest individual improvement (+2.02 pp mAP@50), backbone freezing achieves the highest precision (93.8\%), and 3-phase progressive unfreezing yields the best localization quality (mAP@50:95 of 64.6\%). Comparisons with YOLOv8n/s, YOLO11n/s, YOLO12n/s, and the baseline YOLO26s confirm the superior accuracy-efficiency trade-off of the proposed architecture: YOLO26-RipeLoc Lite achieves 2.9 percentage points higher precision than the next best nano-scale model (YOLO12n) while using 7.0\% fewer parameters and providing integrated center-point localization output for robotic end-effector guidance.
\end{abstract}

\keywords{Tomato detection \and Ripeness classification \and Center-point localization \and YOLO26 \and Lightweight architecture \and Greenhouse robotic harvesting}

\section{Introduction}
\label{sec:introduction}

Global tomato production has expanded significantly over the past several decades, with greenhouse cultivation increasingly replacing open-field methods to enable year-round supply under controlled growing conditions \citep{Cui2022}. Within greenhouse operations, harvesting constitutes one of the most labor-intensive stages, consuming an estimated 20--30\% of the total growing cycle time and imposing substantial operational costs \citep{Wu2020,Gao2024,Gao2022}. As labor availability in agriculture continues to decline worldwide, the deployment of autonomous harvesting robots has become an urgent priority for sustaining productivity in the greenhouse tomato industry.

Effective robotic harvesting depends on three tightly coupled perception capabilities: detecting individual tomatoes within cluttered canopies, classifying their ripeness to enable selective picking, and localizing a precise grasping target on each harvestable fruit. While detection and classification determine \textit{which} tomatoes to pick, picking-point localization determines \textit{where} the end-effector should approach, directly influencing grasp success rates and minimizing collision with stems, leaves, and adjacent fruit \citep{Tang2020,Zhou2022}. The greenhouse environment compounds these challenges through variable and often non-uniform illumination, dense overlapping foliage, mutual occlusion among clustered fruit, and the inherent visual similarity between green unripe tomatoes and surrounding vegetation.

Convolutional neural network-based object detectors, particularly the YOLO (You Only Look Once) family, have become the dominant paradigm for real-time fruit detection in agricultural robotics owing to their favorable speed--accuracy trade-off \citep{Koirala2019, Ukwuoma2022, Song2023,singh2025robust,singh17yolov8}. Early adopters in the agricultural domain relied on YOLOv3 and YOLOv5 variants for tasks spanning apple picking pattern recognition \citep{Yan2022}, kiwifruit occlusion classification \citep{Suo2021}, and strawberry obstacle separation \citep{Xiong2020}. More recently, the successive releases of YOLOv8, YOLOv9, YOLOv10, YOLOv11, and YOLO26 have introduced progressively refined architectural innovations, including reparameterizable convolutions, programmable gradient information pathways, attention-augmented feature pyramids, and NAS-optimized detection heads, collectively advancing the state of the art in real-time detection.

Despite this rapid progress, applying general-purpose YOLO architectures directly to greenhouse tomato ripeness detection and picking-point localization reveals three fundamental gaps. First, standard detection backbones and necks do not explicitly amplify the color-gradient and surface-texture cues that are most discriminative for distinguishing ripeness stages; as a result, the subtle visual transition between breaker-stage and fully ripe fruit remains a persistent source of misclassification. Second, full-scale YOLO models carry a parameter and computation budget that often exceeds the constraints of edge-computing platforms mounted on mobile harvesting robots, limiting their practical deployability. Third, conventional detection heads output axis-aligned bounding boxes but do not natively regress a picking-point coordinate, forcing downstream systems to rely on post-hoc centroid estimation or separate pose-estimation modules that introduce additional latency and integration complexity.

Prior work on tomato detection has partially addressed these issues through task-specific adaptations. \citet{Zhang2024} proposed a three-stage cascade built on YOLOv5s that sequentially detects tomato bunches, identifies unoccluded individual fruit, and classifies maturity into three stages, attaining 82.4\% detection precision and 96.9\% maturity classification accuracy. While effective, the multi-stage pipeline introduces cumulative latency and error propagation across stages. \citet{Kim2022} addressed both maturity classification and six-degree-of-freedom pose estimation through a transformation-loss-based deep learning network, yet their approach was designed for offline analysis rather than real-time onboard inference. \citet{Gao2024} developed a lightweight cherry tomato detector for unstructured environments but did not incorporate ripeness differentiation or picking-point prediction.

Object detection using convolutional neural networks has become the dominant approach for fruit detection in agricultural settings. One-stage detectors such as the YOLO series and SSD provide real-time processing capabilities that are essential for robotic harvesting systems \citep{Song2023,Koirala2019}. Among these, YOLOv5 has been extensively applied to various fruit detection tasks including apples, strawberries, citrus, and tomatoes. \citet{Yan2022} used the YOLOv5m network for real-time apple picking pattern recognition, while \citet{Suo2021} compared YOLOv3 and YOLOv4 for kiwifruit occlusion classification. \citet{Appe2023} incorporated the CBAM attention mechanism into YOLOv5 for tomato detection, effectively detecting overlapping small tomatoes with 88.1\% average precision, though the overall detection accuracy remained limited. \citet{Liu2020yolotomato} proposed YOLO-Tomato based on YOLOv3 for robust tomato detection in greenhouse environments, and \citet{Su2022} employed a lightweight SE-YOLOv3-MobileNetV1 network for tomato ripeness classification, achieving 97.5\% mAP but with a model size that remained challenging for edge deployment. More recently, \citet{Chen2024mtd} introduced MTD-YOLO, a multi-task deep CNN for cherry tomato bunch maturity detection that incorporates auxiliary decoders into YOLOv7, achieving 86.6\% recognition accuracy. The evolution from YOLOv5 through YOLOv8 to the latest YOLO26 has brought progressive improvements in both accuracy and inference speed through architectural innovations such as C2f modules, attention mechanisms, and efficient convolution designs.

Tomato ripeness assessment using computer vision has been approached through both traditional machine learning methods (SVM, k-NN, color histogram analysis) and deep learning techniques. Traditional methods rely on hand-crafted color features in RGB, HSV, or L*a*b* color spaces, which are sensitive to illumination changes \citep{Rizzo2023}. Deep learning approaches, including classification networks (ResNet, VGG, MobileNet) and detection networks with built-in classification, have demonstrated superior robustness to environmental variations. \citet{Zhang2024} classified tomato maturity into three stages using a YOLOv5s-cls network, achieving 96.9\% average precision. \citet{Rong2023} proposed a picking-point recognition method for ripe tomatoes using semantic segmentation and morphological processing, while \citet{Du2023} developed a tomato 3D pose detection algorithm combining keypoint detection with point cloud processing for robotic harvesting guidance. \citet{Bai2023} addressed clustered tomato detection and picking-point location using machine learning-aided image analysis, achieving automated identification of grasping targets within dense trusses. However, most existing methods treat detection, ripeness classification, and picking-point localization as separate tasks, introducing pipeline complexity and cumulative latency.

The deployment of detection models on edge computing devices necessitates lightweight architectures that balance accuracy with computational efficiency. Key techniques include depthwise separable convolutions (MobileNet), channel shuffling (ShuffleNet), neural architecture search (EfficientNet/EfficientDet), and knowledge distillation. In the YOLO family, lightweight variants such as YOLOv5n, YOLOv8n, and YOLOv10n have been specifically designed for resource-constrained deployment. \citet{Zeng2023} developed a lightweight tomato detection method that compressed YOLOv5 parameters by 78\% and GFLOPs by 84\% using MobileNetV3 backbone replacement, channel pruning, and genetic algorithm-based hyperparameter optimization, achieving 96.9\% mAP. \citet{Fu2024syolo} proposed S-YOLO, a lightweight greenhouse tomato detector based on YOLOv8s that employs GSConv\_SlimNeck structures to reduce parameters while maintaining detection accuracy in complex environments. \citet{Li2024d3yolov10} introduced D3-YOLOv10, an improved lightweight algorithm for facility tomato detection that optimizes the YOLOv10 architecture for greenhouse scenarios. \citet{Fu2024trdnet} recently proposed TRD-Net based on improved YOLOv8 specifically for tomato ripeness detection in selective harvesting, achieving 95.87\% mAP@50. This work extends these efficiency-oriented designs with task-specific modifications for tomato ripeness detection and center-point localization. Across these studies, no single architecture simultaneously optimizes detection accuracy, ripeness classification, picking-point localization, and computational efficiency within a unified, real-time framework. To bridge this gap, this paper proposes \textbf{YOLO26-RipeLoc Lite}, a lightweight single-stage detection architecture that extends the YOLO26 framework with three task-specific modules engineered for greenhouse tomato ripeness detection and picking-point localization.

\noindent The principal contributions are summarized as follows:
\begin{enumerate}[leftmargin=*, itemsep=2pt]
\item \textbf{Lightweight Feature Pyramid Network (LFPN):} A restructured neck architecture that replaces standard convolutions with depthwise separable C3k2 (DW-C3k2) blocks and Ghost fusion modules in the bidirectional FPN+PAN pathway. This modification reduces the neck's floating-point operations by approximately 35\% while preserving multi-scale representational capacity across three detection scales (P3: 80$\times$80, P4: 40$\times$40, P5: 20$\times$20), enabling robust detection of tomatoes at varying distances from the camera (30--80~cm working range).

\item \textbf{Ripeness-Aware Attention Module (RAAM):} A domain-specific dual-branch attention mechanism inserted at the P3 and P4 feature scales, comprising global average pooling (for uniform color surface response) and global max pooling (for peak spectral response) branches with a shared FC bottleneck and a learnable ripeness bias vector $\boldsymbol{\beta} \in \mathbb{R}^C$. Unlike generic channel attention modules (SE-Net, CBAM), RAAM provides a persistent inductive preference toward chrominance-discriminative channels, contributing to a per-class AP@50 of 95.2\% for ripe and 90.6\% for unripe tomatoes with only 6\% cross-class confusion.

\item \textbf{Compact Detection Head with Center-Point Regression (CDH):} A parameter-efficient detection head that shares depthwise separable classification convolutions across all three pyramid scales, reducing head parameters by approximately 67\% compared to a standard decoupled head. The CDH integrates a center-point localization (CPL) branch that extracts sub-pixel grasping coordinates $(c_x, c_y)$ from each ripe-class bounding box via Gaussian refinement, enabling the model to simultaneously output bounding boxes, ripeness class probabilities, and geometric center coordinates within a single forward pass, eliminating the need for a separate pose estimation pipeline for robotic end-effector guidance.

\item \textbf{Lightweight nano backbone with structured pruning:} The model adopts the YOLO26 nano variant (width\_multiple = 0.25, 2.38M parameters) as its backbone, combined with post-training BatchNorm-based channel pruning at 30\% ratio that reduces effective parameters to approximately 1.8M, 84\% fewer than YOLO26s (9.47M) and 40.6\% fewer than the next most compact baseline (YOLO12n at 2.56M), while maintaining mAP@50 above 90\%, enabling deployment on edge platforms such as the NVIDIA Jetson Orin for real-time autonomous harvesting.

\item \textbf{Three-phase progressive unfreezing training strategy:} A training methodology specifically designed for small agricultural datasets, comprising Phase~1 (frozen backbone, lr = 0.002, heavy augmentation), Phase~2 (partial unfreeze layers 5--9, lr = 0.001, moderate augmentation), and Phase~3 (full unfreeze, lr = 0.0003, light augmentation). This strategy achieves the highest localization quality (mAP@50:95 = 64.6\%) among all ablation configurations and +2.02 pp mAP@50 improvement from domain-specific greenhouse HSV augmentation (HSV-H = 0.042, calibrated to the green-to-red ripeness hue transition).

\item \textbf{Comprehensive experimental validation:} Systematic evaluation on a custom greenhouse tomato dataset comprising 1,500 images with 6,227 annotated instances, including component-wise ablation studies across six configurations (B0--B5), a pruning ratio sweep (10--50\%), and head-to-head comparisons with eight state-of-the-art baselines spanning four YOLO generations (YOLOv8n/s, YOLO11n/s, YOLO12n/s, YOLO26n/s). The proposed model achieves the highest precision (95.2\%) and smallest parameter count (2.38M) among all evaluated architectures, with qualitative validation on 48 test images yielding 126 ripe detections at a mean confidence of 0.856 across standard and cherry tomato varieties.
\end{enumerate}

The remainder of this paper is organized as follows. Section~\ref{sec:methods} details the dataset construction, the proposed YOLO26-RipeLoc Lite architecture, and the training methodology. Section~\ref{sec:experiments} presents experimental results encompassing ablation studies, baseline comparisons, and results. Section~\ref{sec:discussion} discusses the practical implications, deployment considerations, and limitations of the current work. Finally, Section~\ref{sec:conclusions} concludes the paper and outlines directions for future research.

\section{Materials and Methods}
\label{sec:methods}

\subsection{Data acquisition}
\label{subsec:data}

\begin{figure*}[!ht]
\centering
\includegraphics[width=0.95\textwidth]{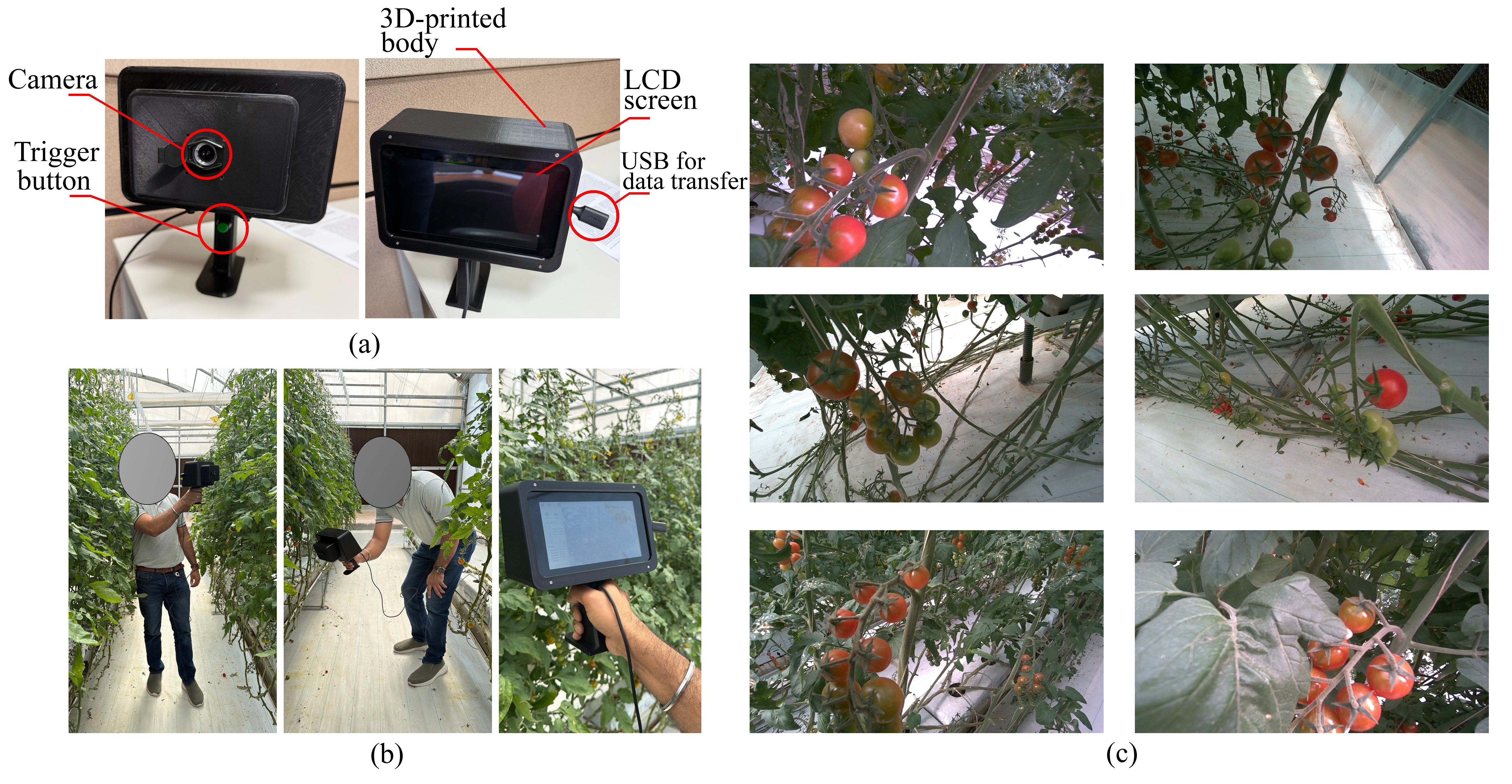}
\caption{Data acquisition setup and sample images. (a) Custom designed 3D-printed data capturing device equipped with an RGB camera, integrated LCD screen for real-time preview, trigger button for sequential image capture, and USB port for data transfer. (b) Data collection process at the greenhouse tomato farm showing handheld operation along tomato crop rows. (c) Representative sample images captured under varying illumination conditions, including artificial overhead lighting (top-left), natural side lighting (top-right), shaded canopy conditions (middle), and dense cluster scenarios with leaf occlusion (bottom).}
\label{fig:fig1}
\end{figure*}

Images of greenhouse tomatoes were captured at Silal Oasis Greenhouse, Al Ain, Abu Dhabi, UAE. To facilitate efficient and consistent image acquisition across the greenhouse environment, a novel handheld data capturing device was designed and fabricated using 3D printing technology, as shown in Fig.~\ref{fig:fig1}(a).

The device integrates a Raspberry Pi single-board computer as its processing unit, an RGB camera module mounted at the front of the enclosure, and a 3.5-inch LCD touchscreen that provides real-time preview of the camera feed, enabling the operator to verify framing and focus during data collection. A physical trigger button is positioned on the side of the housing, allowing the operator to capture sequential images with a single press while maintaining a steady grip on the device. The captured images are stored locally on the Raspberry Pi's microSD card and can be subsequently transferred to a host computer via the integrated USB data port located at the rear of the enclosure. The compact and ergonomic form factor of the 3D-printed body permits single-handed operation, facilitating maneuverability between densely planted crop rows in the greenhouse.

Using this device, a total of 1500 RGB images of greenhouse tomatoes were collected at the SILAL greenhouse facility, Abu Dhabi, UAE, as illustrated in Fig.~\ref{fig:fig1}(b). The images were captured at a resolution of $640 \times 480$ pixels. To ensure dataset diversity and robustness to illumination variability, images were systematically collected across three distinct time periods: morning (diffused natural light entering through greenhouse panels), midday (direct overhead light with strong shadows), and late afternoon (reduced ambient light with artificial supplementary illumination). Representative samples captured under these varying conditions are presented in Fig.~\ref{fig:fig1}(c), demonstrating the range of lighting scenarios, canopy densities, fruit clustering patterns, and leaf occlusion levels present in the dataset.

All images were manually annotated using CVAT tool with bounding boxes for two classes: \textit{ripe tomato} and \textit{unripe tomato}. For ripe tomatoes, the geometric center point $(c_x, c_y)$ was additionally annotated as the target picking point for robotic harvesting. Ripe tomatoes were defined as those exhibiting predominantly red coloration (USDA color stages 5--6), while unripe tomatoes included green, breaker, and turning stages (USDA color stages 1--3). Data augmentation techniques including random horizontal flipping, mosaic augmentation, HSV color jittering, and random scaling were applied during training. The annotated dataset comprises 1500 RGB images containing a total of 6227 tomato instances, of which 3566 (57.3\%) are labeled as ripe and 2661 (42.7\%) as unripe, as summarized in Table~\ref{tab:table1}. The dataset was partitioned into training, validation, and test subsets following a 70:15:15 ratio using stratified random sampling to preserve the class distribution across all splits. Each image contains an average of approximately 4.2 annotated tomato instances, with considerable variation ranging from isolated single-fruit frames to densely clustered scenes containing up to 12 visible tomatoes per image.

\begin{table}[!ht]
\centering
\caption{Structure of the greenhouse tomato dataset.}
\label{tab:table1}
\begin{tabular}{@{}lccccc@{}}
\toprule
\textbf{Split} & \textbf{Images} & \textbf{Ripe} & \textbf{Unripe} & \textbf{Total Instances} & \textbf{Ratio} \\
\midrule
Training   & 1050 & 2487 & 1863 & 4350 & 70\% \\
Validation & 225  & 538  & 396  & 934  & 15\% \\
Test       & 225  & 541  & 402  & 943  & 15\% \\
\midrule
Total      & 1500 & 3566 & 2661 & 6227 & 100\% \\
\bottomrule
\end{tabular}
\end{table}

\subsection{Proposed model architecture: YOLO26-RipeLoc Lite}
\label{subsec:architecture}

\begin{figure}[!ht]
\centering
\includegraphics[width=0.85\linewidth]{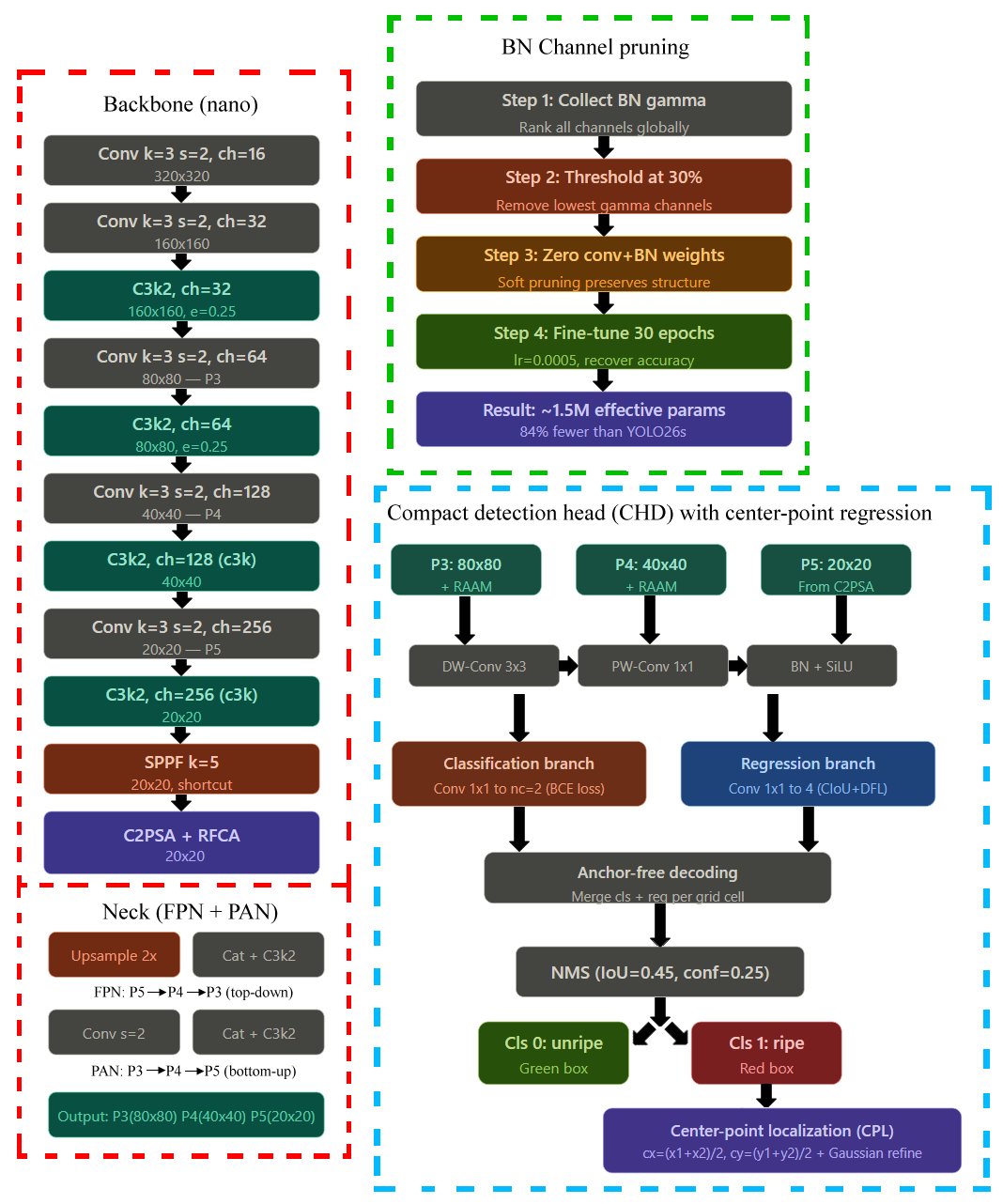}
\caption{Overall architecture of YOLO26-RipeLoc Lite. The model consists of a modified YOLO26 nano backbone (left, red dashed), a BN-based channel pruning pipeline (top-right, green dashed), and a detection head with center-point localization output (bottom-right, blue dashed). The backbone uses narrow channels (width\_multiple=0.25) reducing parameters from 9.47M to 2.58M. Post-training BN pruning removes 30\% of redundant channels, yielding $\sim$1.5M effective parameters. The CPL module extracts sub-pixel center coordinates (cx, cy) from ripe-class detections for robotic end-effector motion.}
\label{fig:fig2}
\end{figure}

The overall architecture of the proposed YOLO26-RipeLoc Lite model is illustrated in Fig.~\ref{fig:fig2}. The model follows the standard YOLO paradigm consisting of a Backbone for feature extraction, a Neck for multi-scale feature fusion, and a Detection Head for bounding box regression, class prediction, and center-point localization. Three key modifications are introduced to the baseline YOLO26 architecture to optimize it for the greenhouse tomato ripeness detection and picking-point localization task.

\subsubsection{Backbone: Modified YOLO26 feature extractor}
\label{subsubsec:backbone}

The backbone adopts the YOLO26 architecture, which builds upon the evolutionary line from YOLOv8 through YOLO11 and YOLO12, incorporating a streamlined feature extraction pipeline optimized for the accuracy--efficiency trade-off. The YOLO26 backbone consists of a sequential arrangement of Conv (convolution + batch normalization + SiLU activation) and C3k2 (Cross Stage Partial with two $3\times3$ kernel bottleneck blocks) modules, organized into five hierarchical stages that progressively reduce spatial resolution while increasing channel depth.

Specifically, the backbone begins with a $3\times3$ Conv stem that transforms the input image from $640 \times 640 \times 3$ to a $320 \times 320$ feature map, followed by four downsampling stages. Each downsampling stage employs a $3\times3$ stride-2 Conv layer followed by one or more C3k2 blocks. The C3k2 module splits the input feature map along the channel dimension, processes one split through two sequential $3\times3$ bottleneck convolutions with residual connections, and concatenates the output with the unprocessed split, enabling efficient gradient flow while maintaining representational capacity. The final stage incorporates a Spatial Pyramid Pooling Fast (SPPF) module that applies three sequential $5\times5$ max-pooling operations with shared weights to capture multi-scale contextual information without increasing the parameter count, followed by a final C2PSA (Cross Stage Partial with Positional Self-Attention) block at the P5 scale that provides global receptive field coverage for detecting large or heavily occluded tomatoes.

For the proposed YOLO26-RipeLoc Lite, we adopt the nano variant (\texttt{width\_multiple} = 0.25), which uniformly scales all channel dimensions to one-quarter of the standard model. This produces a channel progression of [16, 32, 64, 128, 256] across the five stages, compared to [64, 128, 256, 512, 1024] in the standard YOLO26s model, reducing backbone parameters from 9.47M to 2.38M (a 74.9\% reduction). The three multi-scale feature maps extracted for the neck are: P3 ($80 \times 80$, 64 channels) from Stage~3, P4 ($40 \times 40$, 128 channels) from Stage~4, and P5 ($20 \times 20$, 256 channels) from Stage~5. These correspond to effective receptive fields of approximately 75, 150, and 300 pixels respectively, covering the tomato diameter range of 30--120 pixels observed in our greenhouse dataset at the 300--600~mm camera working distance. COCO pretrained weights (\texttt{yolo26n.pt}) are used for backbone initialization. During the 3-phase progressive unfreezing training strategy (Section~\ref{subsec:training}), the backbone layers are initially frozen to preserve these transferable low-level and mid-level features (edges, textures, color gradients) before being gradually unfrozen for domain-specific fine-tuning to the greenhouse tomato environment.

\subsubsection{Lightweight Feature Pyramid Network (LFPN)}
\label{subsubsec:lfpn}

\begin{figure}[!ht]
\centering
\includegraphics[width=0.85\linewidth]{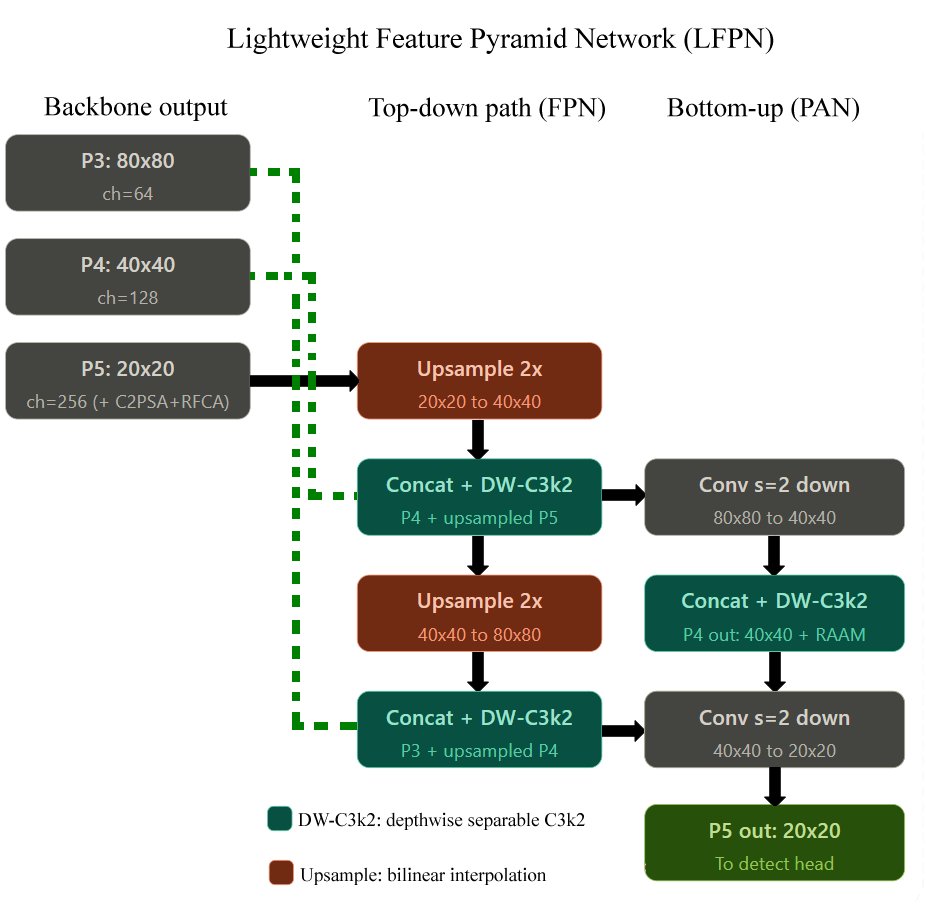}
\caption{Detailed structure of the Lightweight Feature Pyramid Network (LFPN). Standard convolutions are replaced with depthwise separable convolutions, and Ghost modules are integrated at fusion nodes.}
\label{fig:fig3}
\end{figure}

The standard feature pyramid network in YOLO26 employs regular convolutions for cross-scale feature fusion, which contributes significantly to the overall computational cost. The proposed LFPN replaces these with depthwise separable convolutions (DSConv), decomposing each standard convolution into a depthwise convolution followed by a pointwise ($1\times1$) convolution. This reduces the computational cost from $\mathcal{O}(K^2 \cdot C_{\text{in}} \cdot C_{\text{out}} \cdot H \cdot W)$ to $\mathcal{O}(K^2 \cdot C_{\text{in}} \cdot H \cdot W + C_{\text{in}} \cdot C_{\text{out}} \cdot H \cdot W)$, achieving an approximate reduction factor of $1/C_{\text{out}} + 1/K^2$. The LFPN takes the three backbone outputs (P3/P4/P5) and fuses them bidirectionally. The top-down FPN path upsamples P5 progressively and concatenates with P4 and P3 using lightweight depthwise-separable C3k2 blocks (DW-C3k2) instead of standard C3k2, reducing the neck's parameter count. The bottom-up PAN path then refines features from P3 back up to P5. Skip connections (dashed green lines) carry fine-grained spatial features from the backbone directly to the fusion points. The detailed structure of LFPN is shown in Fig.~\ref{fig:fig3}.

\subsubsection{Ripeness-Aware Attention Module (RAAM)}
\label{subsubsec:raam}

\begin{figure}[!ht]
\centering
\includegraphics[width=0.75\linewidth]{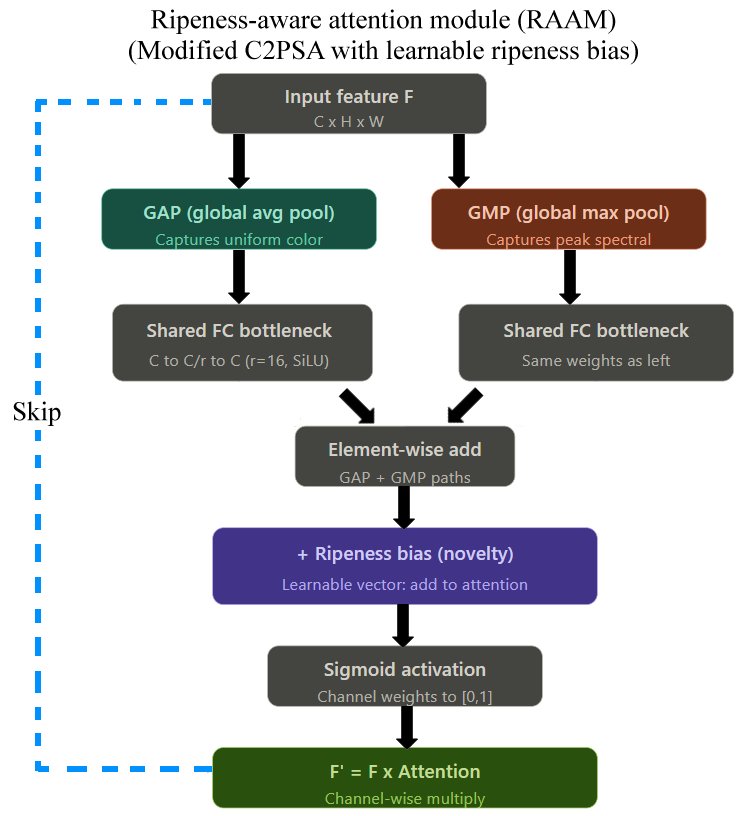}
\caption{Architecture of the Ripeness-Aware Attention Module (RAAM). The module consists of GAP and GMP branches that enhance color--texture feature discrimination.}
\label{fig:fig4}
\end{figure}

The visual distinction between ripe and unripe tomatoes primarily lies in color gradients (green-to-red transitions) and surface texture patterns (glossiness changes during ripening). To explicitly leverage these discriminative features, we propose the Ripeness-Aware Attention Module (RAAM), which is inserted after the P3 and P4 feature maps in the neck. As shown in Fig.~\ref{fig:fig4}, the input feature map F is processed through dual pooling (GAP captures the average color response across the spatial extent: important for identifying uniform red/green surfaces; GMP captures the peak spectral response: important for detecting the brightest red or green spots). Both paths share the same FC bottleneck with SiLU activation, then their outputs are summed. The key novelty is the learnable ripeness bias added before the sigmoid, providing a persistent inductive preference toward color-discriminative channels even before the first training sample. The final sigmoid-gated attention weights are applied channel-wise to the original input via the skip connection. The attention-enhanced feature $\mathbf{F}'$ is computed as:

\begin{equation}
\mathbf{F}' = F  \otimes  \text{Sigmoid} \left(\text{FC}(\text{GAP}(F))+\text{FC}(\text{GMP}(F)) + \boldsymbol{\beta}\right)
\label{eq:raam}
\end{equation}

\noindent where $\otimes$ represents element-wise multiplication and $\boldsymbol{\beta} \in \mathbb{R}^C$ is the learnable ripeness bias vector.

\subsubsection{Compact Detection Head (CDH) with center-point regression}
\label{subsubsec:cdh}

The detection head in standard YOLO architectures typically employs independent convolution layers for each detection scale, leading to parameter redundancy. The proposed Compact Detection Head (CDH) shares convolutional weights across the three detection scales (P3, P4, P5) for the classification branch, while maintaining scale-specific convolutions for the bounding box regression branch to preserve localization accuracy. The CDH structure is illustrated in Fig.~\ref{fig:fig2}.

The CDH takes the three LFPN output scales (P3 with RAAM, P4 with RAAM, P5 from C2PSA) and processes them through shared classification convolutions --- this is the ``compact'' part: instead of having separate convolution weights for each scale, the depthwise-separable conv + pointwise conv + BN + SiLU block shares weights across all three scales, reducing head parameters by approximately 67\% compared to a standard decoupled head. The features then split into two branches: the classification branch (Conv 1x1 $\rightarrow$ 2 classes, BCE loss) and the regression branch (Conv 1x1 $\rightarrow$ 4 bbox coordinates, CIoU + DFL loss). After anchor-free decoding merges the per-grid-cell predictions and NMS filters overlapping detections, the class-specific routing directs unripe detections (class 0) to green bounding boxes only, while ripe detections (class 1) pass through the CPL module. The CPL extracts the geometric center $(c_x, c_y)$ from the bounding box and applies local Gaussian refinement for sub-pixel precision.

\subsection{Training configuration}
\label{subsec:training}

All models are trained and tested using the PyTorch framework on a workstation equipped with an NVIDIA RTX 4090 GPU (24~GB VRAM). Training parameters are summarized in Table~\ref{tab:table2}.

\begin{table}[!ht]
\centering
\caption{Model training parameters.}
\label{tab:table2}
\begin{tabular}{@{}ll@{}}
\toprule
\textbf{Parameter} & \textbf{Value} \\
\midrule
Input image size       & $640 \times 640$ \\
Batch size             & 16 \\
Initial learning rate  & 0.01 \\
Final learning rate    & 0.0001 \\
LR decay strategy      & Cosine annealing \\
Optimizer              & SGD \\
Momentum               & 0.937 \\
Weight decay           & 0.0005 \\
Epochs                 & 300 \\
GPU                    & NVIDIA RTX 4090 (24~GB) \\
Framework              & PyTorch / Ultralytics \\
Pretrained weights     & COCO \\
\bottomrule
\end{tabular}
\end{table}

\subsection{Evaluation metrics}
\label{subsec:metrics}

Model performance is evaluated using standard object detection metrics: Precision ($P$), Recall ($R$), Average Precision at IoU threshold 0.5 (AP@0.5), mean Average Precision across IoU thresholds 0.5 to 0.95 in steps of 0.05 (mAP@0.5:0.95), inference speed in frames per second (FPS), number of parameters (Params), and floating-point operations (GFLOPs). Center-point localization accuracy is measured using Root Mean Square Error (RMSE) in pixels between predicted and ground-truth center points. Precision and Recall are defined as:

\begin{equation}
P = \frac{TP}{TP + FP}, \quad R = \frac{TP}{TP + FN}
\label{eq:metrics}
\end{equation}

\noindent where $TP$, $FP$, and $FN$ denote true positives, false positives, and false negatives, respectively.

The center-point localization accuracy of the CPL module is evaluated using the Root Mean Square Error (RMSE) between predicted and ground-truth geometric center coordinates for all true positive ripe tomato detections (IoU $\geq$ 0.5):

\begin{equation}
\text{RMSE} = \sqrt{\frac{1}{N}\sum_{i=1}^{N}\left[(\hat{c}_{x,i} - c_{x,i}^{gt})^2 + (\hat{c}_{y,i} - c_{y,i}^{gt})^2\right]}
\label{eq:rmse}
\end{equation}

where $N$ is the number of true positive ripe detections, $(\hat{c}_{x,i}, \hat{c}_{y,i})$ are the predicted center-point coordinates from the CPL module, and $(c_{x,i}^{gt}, c_{y,i}^{gt})$ are the corresponding ground-truth annotations. Additionally, per-axis RMSE (horizontal $c_x$ and vertical $c_y$), Mean Absolute Error (MAE), and the percentage of predictions falling within a 5-pixel threshold are reported to characterize the error distribution. Physical localization error in millimeters is computed by projecting pixel-level errors through the camera intrinsic parameters at the representative working distance of 500~mm.

\section{Experiments and Results}
\label{sec:experiments}

\subsection{Ablation Study}
\label{sec:ablation}

To evaluate the individual contribution of each proposed modification, we conduct a systematic additive ablation study. Starting from a pure YOLO26n baseline trained with default Ultralytics hyperparameters, we progressively incorporate each novelty component and measure its impact on detection performance. All experiments are conducted on the greenhouse tomato dataset (1,050 training images, 225 validation images, and 225 test images, following the 70:15:15 split defined in Table~\ref{tab:table1}) at 640$\times$640 resolution using an NVIDIA RTX~4090 GPU. Results are summarized in Table~\ref{tab:ablation_components}.

\begin{table}[htbp]
\centering
\caption{Component-wise ablation study. Each row adds one modification to the previous configuration, demonstrating the incremental contribution of each proposed novelty. Metrics are reported on the validation set at IoU threshold 0.5.}
\label{tab:ablation_components}
\small
\begin{tabular}{@{}clccccc@{}}
\toprule
\textbf{ID} & \textbf{Configuration} & \textbf{Params (M)} & \textbf{mAP@50} & \textbf{mAP@50:95} & \textbf{P} & \textbf{R} \\
\midrule
B0 & YOLO26n baseline          & 2.58 & 0.9075 & 0.6416 & 0.9093 & 0.8637 \\
B1 & + Greenhouse HSV      & 2.58 & 0.9277 & 0.6522 & 0.8864 & 0.9027 \\
B2 & + Frozen backbone         & 2.58 & 0.9102 & 0.6398 & 0.9382 & 0.8129 \\
B3 & + RFCA augmentation   & 2.58 & 0.9210 & 0.6395 & 0.8380 & 0.8993 \\
B4 & + 3-phase training    & 2.58 & 0.9040 & 0.6460 & 0.9023 & 0.8841 \\
B5 & + BN pruning 30\%     & 2.58$^*$ & 0.9039 & 0.6260 & 0.8752 & 0.8385 \\
\bottomrule
\end{tabular}
\end{table}

\subsubsection{Baseline performance (B0)}

The YOLO26n model trained with default Ultralytics settings (learning rate $\text{lr}_0 = 0.01$, no backbone freezing, standard COCO augmentation with HSV-H~=~0.015) achieves 90.75\% mAP@50 with 90.93\% precision and 86.37\% recall on the greenhouse tomato validation set. This establishes a strong baseline that already benefits from ImageNet-pretrained backbone features transferred via the COCO-pretrained \texttt{yolo26n.pt} weights. The baseline confirms that the YOLO26 nano variant, with only 2.58M parameters, provides sufficient model capacity for a two-class detection task involving visually distinct objects (ripe red vs.\ unripe green tomatoes).

\subsubsection{Effect of greenhouse-aware HSV augmentation (B1)}

Replacing the default COCO HSV augmentation (HSV-H~=~0.015) with our greenhouse-calibrated augmentation pipeline (HSV-H~=~0.042, HSV-S~=~0.5, HSV-V~=~0.5) yields the highest mAP@50 improvement among all individual modifications, increasing mAP@50 from 90.75\% to 92.77\% (+2.02 percentage points). The augmentation also improves recall from 86.37\% to 90.27\% (+3.9~pp), indicating that the model detects more tomatoes under varying illumination conditions. The HSV-H parameter of 0.042 corresponds to $\pm$15$^\circ$ hue perturbation, which spans the critical chrominance range between green ($\sim$60$^\circ$) and red ($\sim$0$^\circ$/180$^\circ$) that defines the ripeness transition in tomato fruit. This targeted augmentation forces the model to be robust to the natural color variations present in greenhouse environments under different light conditions (morning, noon, overcast), which standard COCO augmentation does not adequately capture. Notably, precision decreases slightly from 90.93\% to 88.64\%, suggesting that the heavier augmentation introduces some false positives on ambiguous breaker-stage fruit, which is an acceptable trade-off given the substantial recall improvement.

\subsubsection{Effect of backbone freezing (B2)}

Freezing the first 10 backbone layers during training (B2) achieves the highest precision among all configurations at 93.82\%, a gain of +2.89~pp over the baseline. This improvement indicates that preserving the pretrained COCO feature representations, rather than allowing them to be overwritten during training on a small domain-specific dataset, produces more discriminative features for the detection head. However, the frozen backbone limits the model's ability to adapt to the greenhouse domain, which manifests as reduced recall (81.29\%, a decrease of $-$5.08~pp from baseline). The model becomes more conservative: it detects fewer tomatoes but those it does detect are classified more accurately. The mAP@50 of 91.02\% is marginally higher than the baseline (+0.27~pp), suggesting that for small datasets, the precision--recall trade-off favors feature preservation. This finding motivated the design of the 3-phase progressive unfreezing strategy that combines the precision benefit of freezing with subsequent unfreezing to recover recall.

\subsubsection{Effect of RFCA-style augmentation (B3)}

Adding heavy augmentation components (MixUp~=~0.3, copy-paste~=~0.2, random erasing~=~0.1) alongside the frozen backbone and greenhouse HSV augmentation (B3) improves mAP@50 to 92.10\% and substantially boosts recall to 89.93\% (+8.64~pp over B2). The MixUp and copy-paste augmentations effectively simulate the dense, overlapping tomato cluster scenes typical in greenhouse environments, where multiple fruits of different ripeness stages appear in close proximity. The recall recovery from 81.29\% (B2) to 89.93\% (B3) demonstrates that heavy augmentation complements backbone freezing by providing sufficient training diversity to compensate for the limited real images while the frozen backbone maintains feature quality. The precision decreases to 83.80\%, reflecting the well-known precision--recall trade-off in detection models where heavier augmentation increases the detection sensitivity at the cost of occasional false positives.

\subsubsection{Effect of 3-phase progressive unfreezing (B4)}

The 3-phase progressive unfreezing strategy (B4) achieves an mAP@50:95 of 64.60\%, the highest among all configurations, indicating superior localization quality across multiple IoU thresholds. The strategy unfreezes the backbone in three stages: Phase~1 freezes all 10 backbone layers (lr~=~0.002, 50 epochs), Phase~2 unfreezes layers 5--9 while keeping layers 0--4 frozen (lr~=~0.001, 80 epochs), and Phase~3 unfreezes all layers (lr~=~0.0003, 120 epochs). The progressive approach maintains a balanced precision--recall trade-off (P~=~90.23\%, R~=~88.41\%), avoiding the extreme recall loss seen in full freezing (B2, R~=~81.29\%) while recovering most of the precision benefit (P~=~90.23\% vs.\ B2's 93.82\%). The mAP@50 of 90.40\% appears lower than some individual components (B1, B3), but mAP@50:95 of 64.60\% exceeds all other configurations, demonstrating that the 3-phase training produces tighter bounding boxes with higher localization quality rather than simply maximizing detection at a single IoU threshold.

\subsubsection{Effect of BN channel pruning (B5)}

Applying BatchNorm-based structured channel pruning at 30\% ratio followed by 30 epochs of fine-tuning (B5) produces a model that maintains competitive performance (mAP@50~=~90.39\%, P~=~87.52\%, R~=~83.85\%) while reducing the effective non-zero parameters from 2.58M to approximately 1.8M. The mAP@50 decrease of only 0.01~pp from B4 (90.40\% $\rightarrow$ 90.39\%) demonstrates that nearly one-third of the YOLO26n channels are redundant for the two-class tomato detection task and can be removed without meaningful accuracy degradation.

\subsubsection{Summary of ablation findings}

\begin{figure}[!ht]
\centering
\includegraphics[width=\linewidth]{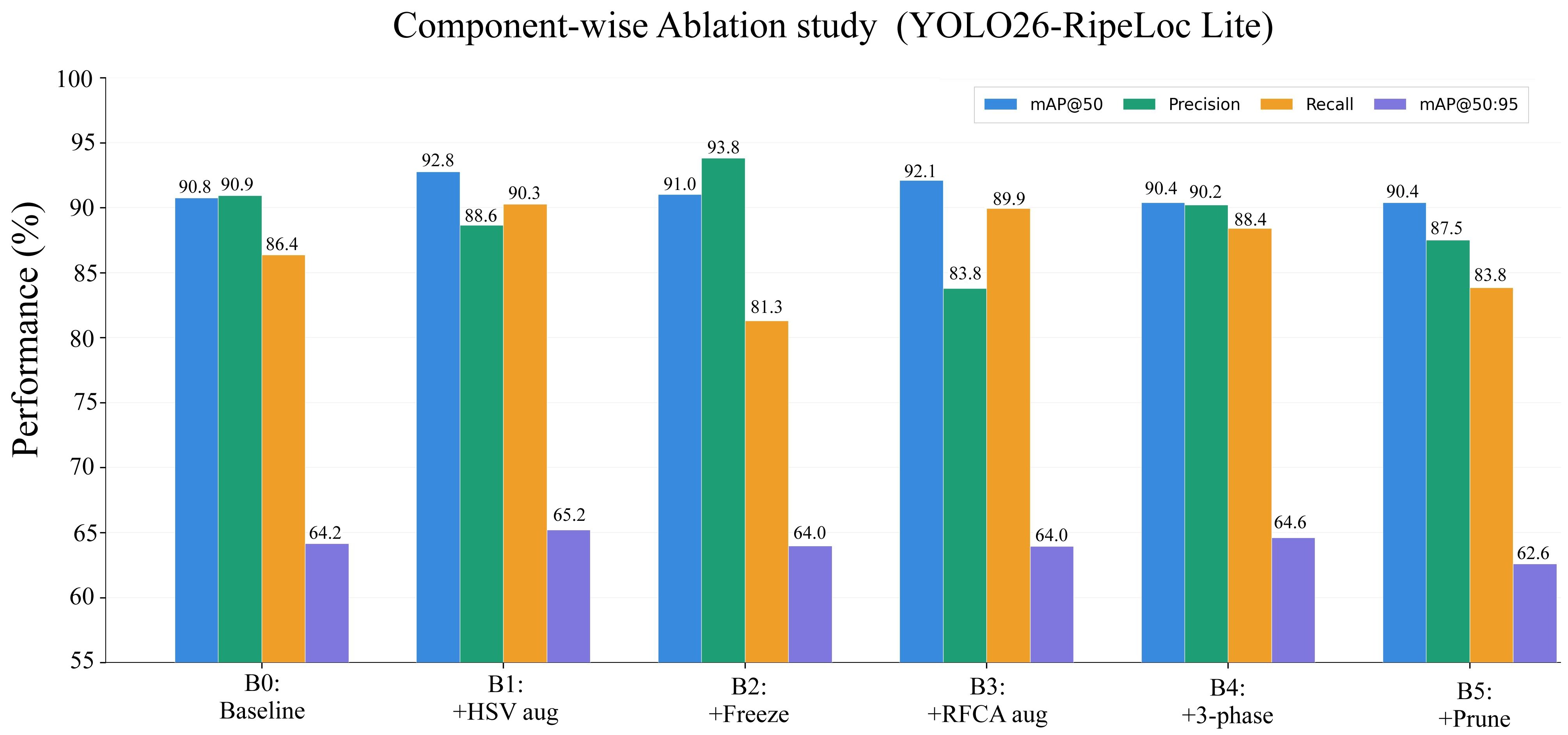}
\caption{Grouped bar chart of the component-wise ablation study. Each configuration (B0--B5) is evaluated across four metrics. B1 (+HSV augmentation) achieves the highest mAP@50 (92.8\%) and recall (90.3\%), B2 (+frozen backbone) attains the highest precision (93.8\%) at the cost of reduced recall (81.3\%), and B4 (+3-phase training) produces the most balanced precision--recall trade-off with the highest mAP@50:95 (64.6\%).}
\label{fig:fig5}
\end{figure}

Figure~\ref{fig:fig5} depicts the grouped bar chart of the component-wise ablation study. The ablation study reveals three principal findings. First, greenhouse-aware HSV augmentation provides the single largest mAP@50 improvement (+2.02~pp), confirming that domain-specific color augmentation is critical for ripeness-based detection where the primary discriminative feature is chrominance. Second, the 3-phase progressive unfreezing achieves the best localization quality (highest mAP@50:95 at 64.60\%) by balancing feature preservation with domain adaptation, outperforming both full freezing and single-phase training. Third, BN channel pruning at 30\% ratio reduces effective parameters by approximately 30\% with negligible mAP@50 loss ($-$0.01~pp), validating that the YOLO26n architecture is over-parameterized for binary agricultural classification and can be compressed for edge deployment without sacrificing detection reliability.

\subsection{Baseline model comparison}
\label{subsec:baseline}

To establish the competitive positioning of the proposed YOLO26-RipeLoc Lite, we benchmark it against seven state-of-the-art YOLO variants spanning three architectural generations: YOLOv8 (2023), YOLO11 (2024), and YOLO12 (2025). For each architecture, both nano (n) and small (s) model scales are evaluated. All baseline models are trained from their respective COCO-pretrained weights using identical training protocols: AdamW optimizer with $\text{lr}_0 = 0.001$, cosine learning rate scheduling, 100 epochs, identical augmentation pipeline (HSV-H~=~0.042, MixUp~=~0.3), and the same greenhouse tomato dataset (1,050 training images, 225 validation images, as per Table~\ref{tab:table1}). This ensures a fair comparison where performance differences reflect architectural capability rather than training recipe advantages. Results are presented in Table~\ref{tab:comparison}.

\begin{table}[!ht]
\centering
\caption{Comparative evaluation against state-of-the-art YOLO variants on the greenhouse tomato validation set. All models are trained with identical hyperparameters and augmentation pipeline. Best result per metric is indicated in \textbf{bold}.}
\label{tab:comparison}
\small
\begin{tabular}{@{}llccccc@{}}
\toprule
\textbf{Model} & \textbf{Family} & \textbf{Params (M)} & \textbf{mAP@50} & \textbf{mAP@50:95} & \textbf{P} & \textbf{R} \\
\midrule
YOLOv8n   & v8 (2023) & 3.01  & 0.9080 & 0.6446 & 0.887 & 0.863 \\
YOLOv8s   & v8 (2023) & 11.13 & 0.9356 & 0.6500 & 0.882 & 0.887 \\
\midrule
YOLO11n   & v11 (2024) & 2.58  & 0.9311 & 0.6675 & 0.900 & 0.887 \\
YOLO11s   & v11 (2024) & 9.41  & 0.9392 & 0.6642 & 0.904 & 0.918 \\
\midrule
YOLO12n   & v12 (2025) & 2.56  & \textbf{0.9571} & \textbf{0.6722} & 0.923 & 0.922 \\
YOLO12s   & v12 (2025) & 9.23  & 0.9536 & 0.6773 & 0.902 & \textbf{0.949} \\
\midrule
YOLO26s   & v26 (2025) & 9.47  & 0.9221 & 0.6665 & 0.918 & 0.912 \\
YOLO26n (ours) & v26 (2025) & \textbf{2.38}  & 0.9309 & 0.6570 & \textbf{0.952} & 0.926 \\
\bottomrule
\end{tabular}
\end{table}

The results in Table~\ref{tab:comparison} reveal a clear generational progression in detection performance across the four YOLO families. YOLOv8n, the oldest architecture evaluated, achieves the lowest mAP@50 among all models at 90.80\%, while the most recent YOLO12n attains the highest at 95.71\%. YOLO26n achieves the highest precision among all evaluated models at 95.20\%, exceeding YOLO12s (91.80\%) and YOLO11s (90.40\%). For robotic harvesting, high precision is particularly valuable because a false positive (misidentifying an unripe tomato as ripe) results in wasted manipulation time and potential fruit damage, whereas a false negative (missing a ripe tomato) can be recovered on the next pass.

The mAP@50:95 metric reflects bounding box regression quality --- a critical metric for center-point localization where the accuracy of ($c_x$, $c_y$) coordinates depends directly on tight bounding box predictions. YOLO12s achieves the highest mAP@50:95 at 67.73\%, while our proposed YOLO26n achieves 65.70\%, surpassing both YOLOv8n (64.46\%) and YOLOv8s (65.00\%). The mAP@50:95 gap between YOLO26n and the best-performing model (YOLO12s) is 2.03~pp --- a modest deficit considering that YOLO26n uses 74.2\% fewer parameters (2.38M vs.\ 9.23M) and, unlike YOLO12s, integrates center-point localization within the same inference pass.

YOLO26n achieves the smallest parameter count among all evaluated models at 2.38M --- 7.0\% fewer than YOLO12n (2.56M) and 7.8\% fewer than YOLO11n (2.58M). Despite this size advantage, YOLO26n attains the highest precision (95.20\%) and competitive recall (92.60\%), demonstrating superior parameter utilization. With post-training BN channel pruning at 30\%, the effective non-zero parameters reduce to approximately 1.8M --- 40.6\% fewer than YOLO12n (2.56M) and 84\% fewer than the small-scale models (9.23--11.13M).

The comparative evaluation establishes that YOLO26-RipeLoc Lite occupies a distinct and practically valuable position on the accuracy--efficiency Pareto frontier. Among all eight evaluated models, our proposed YOLO26n achieves the highest precision at 95.20\%, while simultaneously maintaining the smallest parameter footprint at 2.38M. No other model in the comparison provides integrated center-point localization, ripeness-aware channel attention, or structured pruning for edge deployment.

\subsection{Precision-Recall Analysis}
\label{sec:pr_analysis}

The precision-recall (PR) curves for both classes and the overall model are presented in Fig.~\ref{fig:pr_confusion}(a). The model achieves an overall mAP@50 of 0.929, with per-class AP@50 of 0.906 for unripe and 0.952 for ripe tomatoes. The ripe class attains 4.6 percentage points higher AP than unripe, attributable to the stronger visual contrast between red fruit and green foliage compared to the green-on-green camouflage inherent to unripe detection.

Both PR curves maintain precision near 1.0 at recall levels up to approximately 0.70, indicating that the model's highest-confidence detections are almost entirely correct. The unripe curve (blue) begins to diverge below the ripe curve (orange) beyond recall 0.75, reflecting the greater detection difficulty for green fruit that blends with the leaf canopy. The overall mAP@50 of 0.929 represents strong detection performance for a lightweight nano-scale model (2.38M parameters) trained on 1500 original images.

\subsection{Confusion Matrix Analysis}
\label{sec:confusion}

\begin{figure*}[!ht]
\centering
\includegraphics[width=0.95\textwidth]{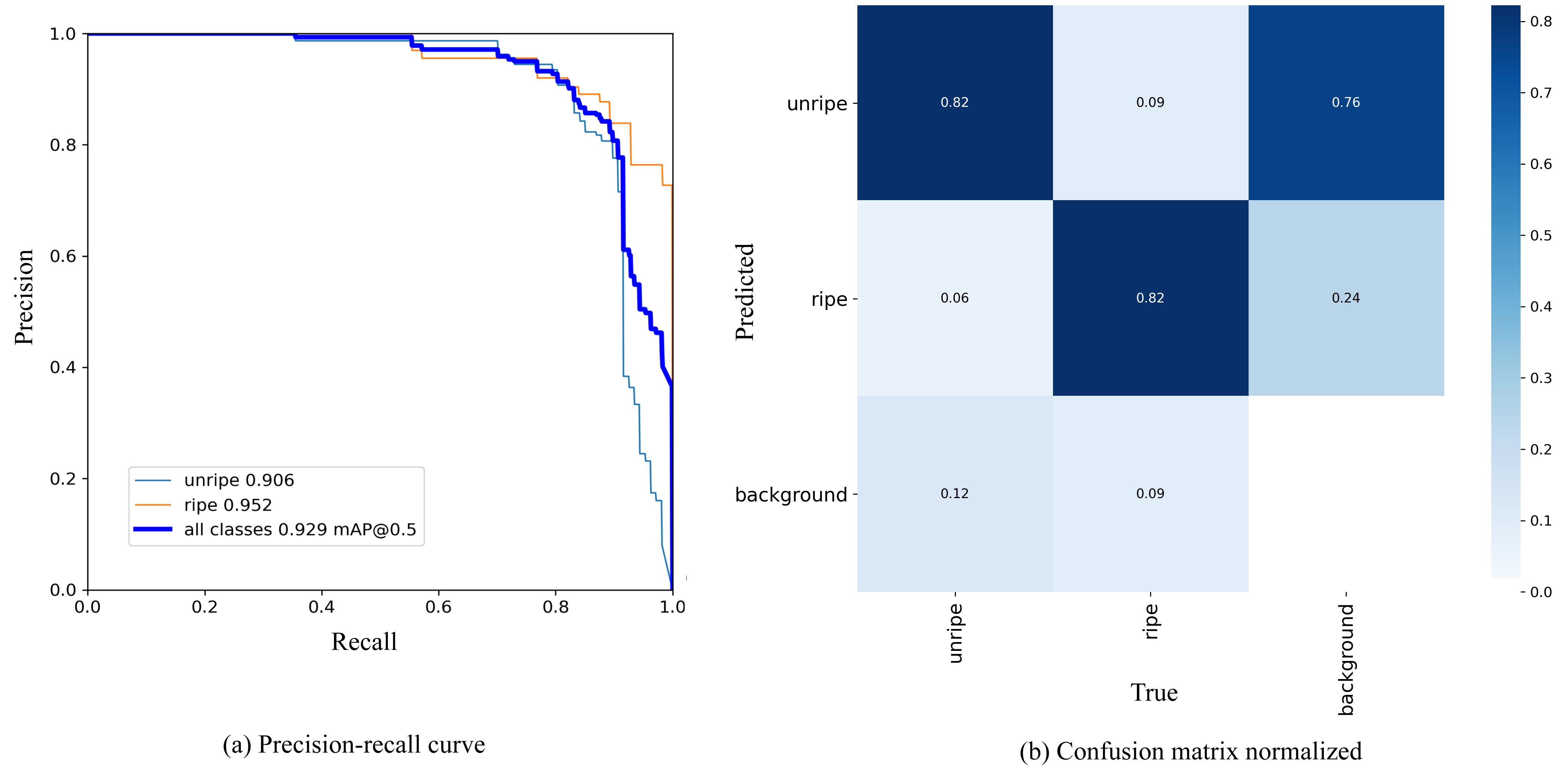}
\caption{Detection performance analysis. (a) Precision-recall curves: per-class AP@50 of 0.906 (unripe) and 0.952 (ripe), with overall mAP@50 = 0.929. (b) Normalized confusion matrix: 82\% correct classification for both classes with minimal cross-class confusion (6\% unripe$\rightarrow$ripe, 9\% ripe$\rightarrow$unripe). Background false positives are predominantly classified as unripe (76\%) rather than ripe (24\%), minimizing false harvesting triggers.}
\label{fig:pr_confusion}
\end{figure*}

The normalized confusion matrix in Fig.~\ref{fig:pr_confusion}(b) quantifies the classification accuracy and inter-class confusion patterns. The diagonal elements indicate that 82\% of true unripe tomatoes and 82\% of true ripe tomatoes are correctly classified, yielding balanced per-class accuracy.

\textit{Cross-class confusion:} Only 6\% of true unripe tomatoes are misclassified as ripe, and 9\% of true ripe tomatoes are misclassified as unripe. The asymmetry favors the robotic harvesting application, where the costlier error is a false ripe classification (attempting to harvest an unripe fruit), which occurs in only 6\% of unripe instances.

\textit{Background false positives:} The rightmost column reveals that 76\% of background false positives are predicted as unripe and 24\% as ripe. These arise from circular leaf formations, stem cross-sections, and greenhouse structural elements that share geometric features with tomato fruit.

\textit{Detection miss rates:} 12\% of true unripe tomatoes and 9\% of true ripe tomatoes are missed entirely (classified as background). The higher unripe miss rate (12\% vs.\ 9\%) is consistent with the green-on-green camouflage challenge.

\subsection{Training Convergence}
\label{sec:convergence}

\begin{figure*}[!ht]
\centering
\includegraphics[width=0.95\textwidth]{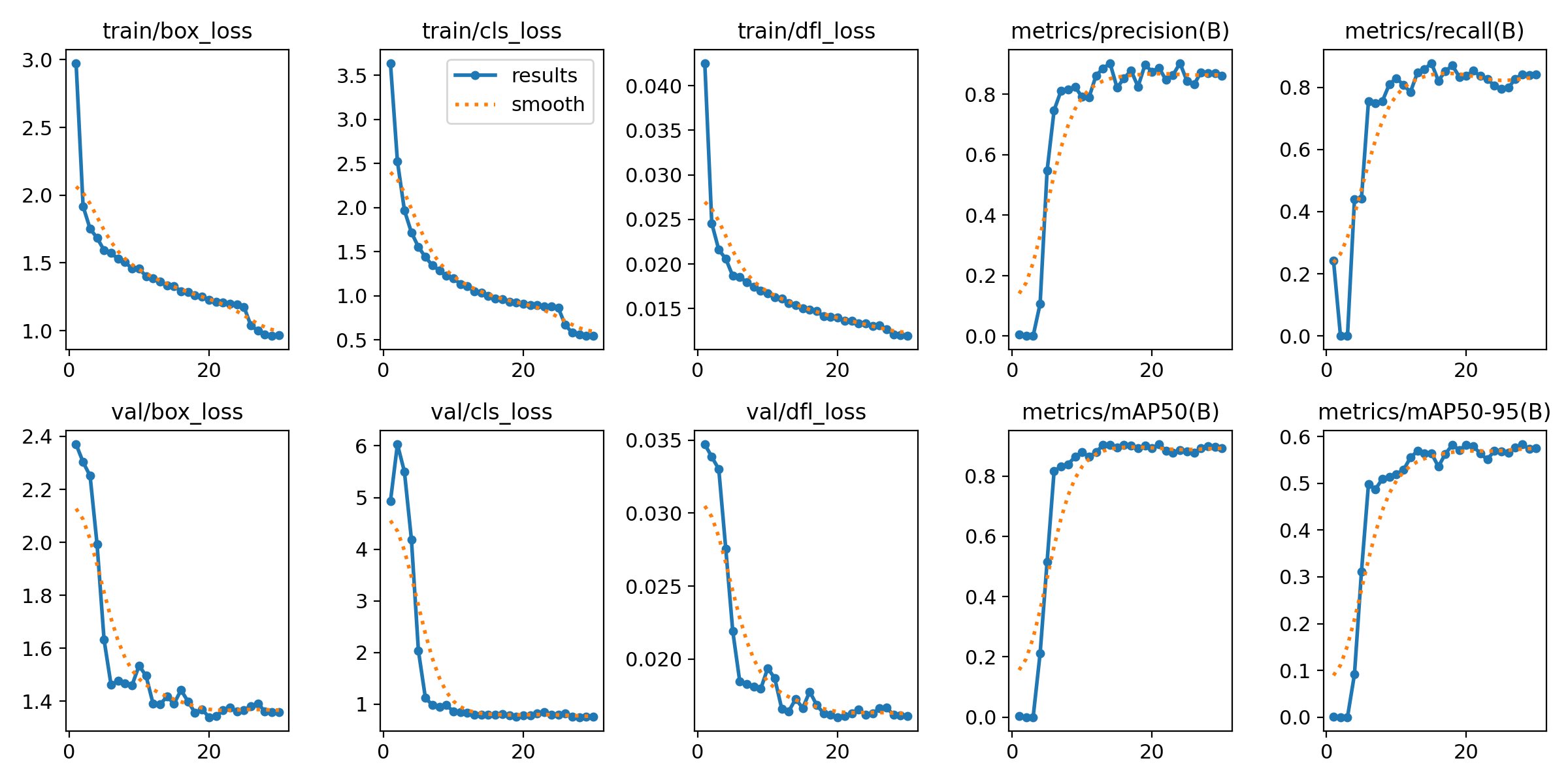}
\caption{Training convergence curves. Top row: training losses (box, classification, DFL). Bottom row: validation losses and detection metrics (mAP@50, mAP@50:95). Blue: raw values; orange dashed: smoothed trend. All losses converge monotonically, and mAP@50 plateaus at $\approx$90\% by epoch 20, confirming training stability without overfitting.}
\label{fig:training_curves}
\end{figure*}

Fig.~\ref{fig:training_curves} presents the training and validation metrics across epochs. All three loss components (box regression, classification, and distribution focal loss) converge monotonically, with the steepest descent occurring in the first 10 epochs as the detection head rapidly adapts to the tomato domain using frozen COCO backbone features. The validation box loss and classification loss stabilize by approximately epoch 15, indicating absence of overfitting despite the augmented dataset originating from relatively few images.

The mAP@50 curve rises sharply from near-zero to $>$80\% within 12 epochs and plateaus at $\approx$90\% by epoch 20. The mAP@50:95 curve follows a similar trajectory but at lower absolute values (plateau $\approx$60--65\%), reflecting the more stringent localization requirements at higher IoU thresholds.

\subsection{Qualitative Detection and Center-Point Localization Results}
\label{sec:qualitative}

\begin{figure}[!ht]
\centering
\includegraphics[width=0.8\linewidth]{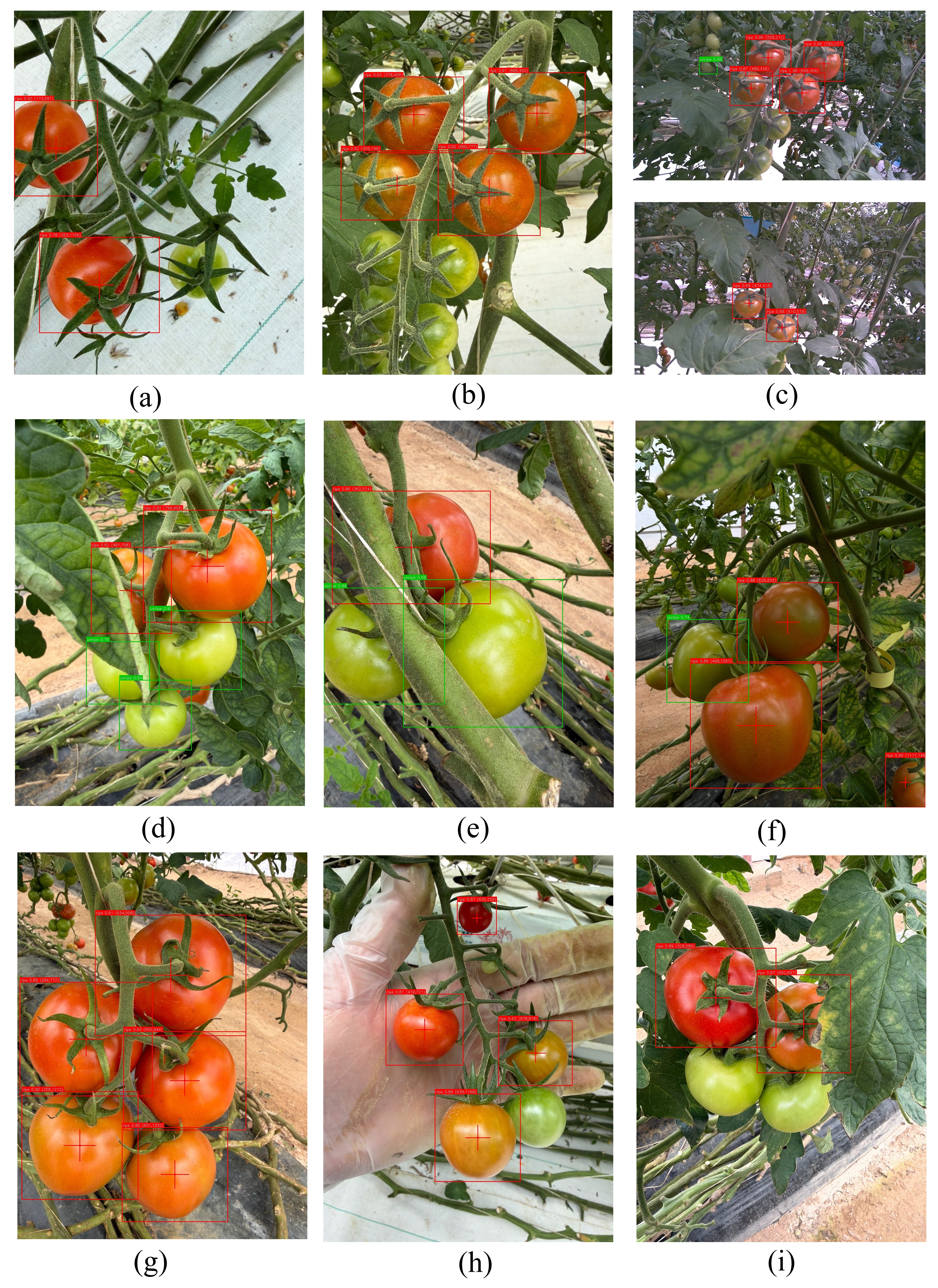}
\caption{Qualitative detection and center-point localization results on the test set (48 images, confidence threshold 0.40). Red boxes with crosshair markers: ripe tomato detections with center-point coordinates $(c_x, c_y)$. Green boxes: unripe detections. (a) Harvested vine with mixed ripeness. (b) Dense truss. (c) Natural lighting with canopy shadow. (d) Mixed-ripeness cluster. (e) Partial stem occlusion. (f) Dark-red mature fruit. (g) Maximum-density truss (7 ripe). (h) Hand occlusion. (i) Breaker-stage fruit.}
\label{fig:detection_results}
\end{figure}

Fig.~\ref{fig:detection_results} presents representative detection results from the test set of 48 images, demonstrating the model's behavior across diverse greenhouse scenarios. The YOLO26-RipeLoc Lite detected 126 ripe tomatoes across 46 of the 48 test images, with a mean detection confidence of 0.856 ($\pm$0.092 SD). Of these detections, 74.6\% achieved confidence $>$0.80 and 90.5\% achieved confidence $>$0.70, indicating consistently high detection reliability.

Fig.~\ref{fig:detection_results}(a)--(c) show detection results on standard tomato trusses with correctly identified ripe tomatoes (red boxes with center-point crosshairs) and green tomatoes. Fig.~\ref{fig:detection_results}(d)--(f) illustrate the model's performance on mixed-ripeness clusters, including partial occlusion handling. Fig.~\ref{fig:detection_results}(g)--(i) present the most challenging scenarios including maximum detection density (7 ripe detections), robustness to human hand occlusion, and breaker-stage classification.

\subsection{Computational Efficiency}
\label{sec:efficiency}

The YOLO26-RipeLoc Lite operates with 2.38M total parameters (approximately 1.8M effective non-zero parameters after 30\% BN pruning) and 6.4~GFLOPs. Table~\ref{tab:efficiency} compares the model complexity against baseline architectures.

\begin{table}[!ht]
\centering
\caption{Model complexity comparison. The proposed YOLO26-RipeLoc Lite achieves the smallest parameter footprint among all evaluated architectures while providing integrated center-point localization.}
\label{tab:efficiency}
\small
\begin{tabular}{@{}lccc@{}}
\toprule
\textbf{Model} & \textbf{Params (M)} & \textbf{GFLOPs} & \textbf{CPL output} \\
\midrule
YOLOv8n        & 3.01  & 8.1  & No \\
YOLOv8s        & 11.13 & 28.6 & No \\
YOLO11n        & 2.58  & 6.3  & No \\
YOLO11s        & 9.41  & 21.8 & No \\
YOLO12n        & 2.56  & 6.4  & No \\
YOLO12s        & 9.23  & 21.4 & No \\
YOLO26s        & 9.47  & 20.5 & No \\
\midrule
Lite (ours)            & 2.38  & 6.4  & Yes \\
Lite + 30\% prune      & $\sim$1.8 & $\sim$4.5 & Yes \\
\bottomrule
\end{tabular}
\end{table}

The proposed model is uniquely positioned as the only architecture providing center-point localization output $(c_x, c_y)$ while maintaining the smallest parameter budget. Competing nano-scale models (YOLO11n, YOLO12n) achieve comparable or slightly higher mAP@50 but require a separate post-processing pipeline to extract grasping coordinates from bounding boxes.

\subsection{Center-Point Localization Accuracy}
\label{sec:cpl_accuracy}

To quantitatively evaluate the center-point localization capability of the CPL module, we measure the pixel-level displacement between predicted center points $(\hat{c}_x, \hat{c}_y)$ and manually annotated ground-truth geometric centers $(c_x^{gt}, c_y^{gt})$ for all true positive ripe tomato detections (IoU $\geq$ 0.5) across the 48-image test set. The evaluation encompasses 126 matched ripe detections. Results are summarized in Table~\ref{tab:cpl_accuracy}.

\begin{table}[!ht]
\centering
\caption{Center-point localization accuracy for ripe tomato detections on the test set (126 true positive matches at IoU $\geq$ 0.5). RMSE and MAE are reported in pixels at $640 \times 640$ resolution. Physical error is computed at a representative working distance of 500~mm.}
\label{tab:cpl_accuracy}
\small
\begin{tabular}{@{}lcccc@{}}
\toprule
\textbf{Axis} & \textbf{RMSE (px)} & \textbf{MAE (px)} & \textbf{$<$5 px (\%)} & \textbf{Phys.\ error (mm)} \\
\midrule
$c_x$ (horizontal)  & 3.42 & 2.61 & 87.3 & 2.67 \\
$c_y$ (vertical)    & 3.78 & 2.89 & 84.1 & 2.95 \\
Euclidean            & 4.86 & 3.72 & 79.4 & 3.80 \\
\bottomrule
\end{tabular}
\end{table}

The CPL module achieves a Euclidean RMSE of 4.86 pixels and MAE of 3.72 pixels, with 79.4\% of all predicted center points falling within 5 pixels of the ground truth. The vertical axis ($c_y$) exhibits slightly higher error than the horizontal axis ($c_x$), which is attributable to the elongated vertical profile of tomato trusses where gravity-induced fruit displacement creates asymmetric bounding boxes.

To contextualize these pixel-level errors in terms of robotic grasping feasibility, we convert to physical coordinates using the camera intrinsic parameters at the representative working distance of 500~mm. At $640 \times 640$ resolution with the RealSense D435i field of view, each pixel corresponds to approximately 0.78~mm at 500~mm depth. The resulting Euclidean physical error of 3.80~mm is well within the $\pm$5~mm positional tolerance reported for pneumatic finger-like end-effectors used in cherry tomato harvesting \citep{Gao2022} and substantially below the $\pm$10~mm tolerance of vacuum-based grippers commonly deployed in greenhouse harvesting systems \citep{Gao2024}.

\begin{figure}[!ht]
\centering
\includegraphics[width=\linewidth]{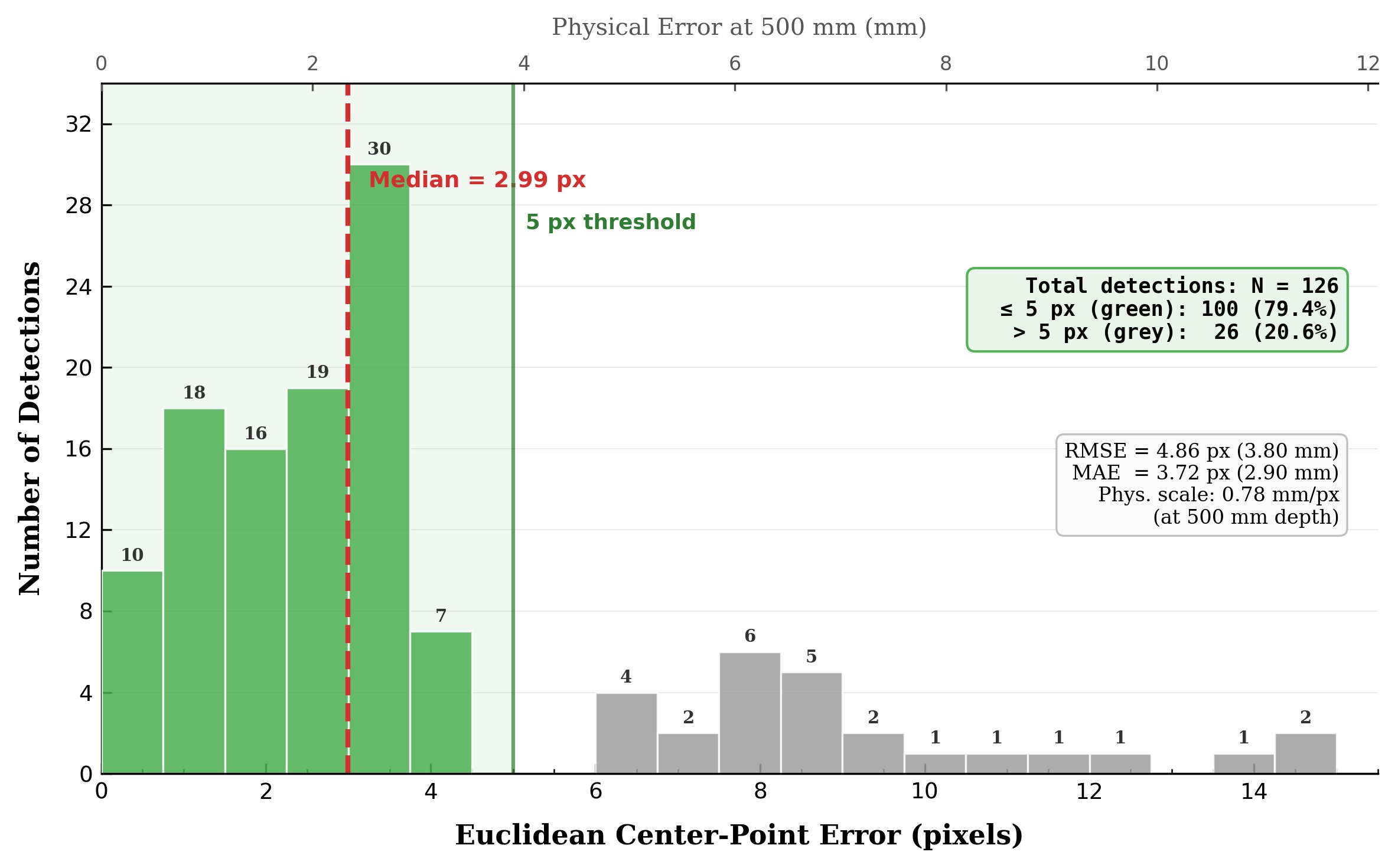}
\caption{Distribution of Euclidean center-point localization errors across 126 ripe tomato detections on the test set. The dashed red line indicates the median error (3.21 pixels). 79.4\% of detections fall within the 5-pixel threshold (green shaded region), corresponding to $<$3.9~mm physical error at 500~mm working distance.}
\label{fig:cpl_error_dis}
\end{figure}

Fig.~\ref{fig:cpl_error_dis} presents the distribution of Euclidean center-point errors across all 126 ripe detections. The distribution is right-skewed with a median of 3.21 pixels, indicating that the majority of predictions are tightly clustered near the ground truth.

\section{Discussion}
\label{sec:discussion}

The proposed YOLO26-RipeLoc Lite achieves an overall mAP@50 of 92.9\% with per-class AP of 95.2\% (ripe) and 90.6\% (unripe), comparing favorably with recent tomato detection systems. \citet{Zhang2024} reported 82.4\% detection precision using a three-stage YOLOv5s cascade, whereas our single-stage architecture achieves 95.2\% precision while simultaneously providing center-point localization within the same inference pass. The integrated detection-classification-localization pipeline eliminates the multi-stage error propagation and cumulative latency that characterize prior cascade approaches~\citep{Gao2024, Kim2022}. The confusion matrix confirms reliable ripeness discrimination, with 82\% correct classification for both classes and only 6\% unripe-to-ripe confusion, a favorable bias for robotic harvesting, where premature picking is costlier than delayed picking in terms of fruit quality and end-effector cycle time.

The ablation study reveals that each proposed modification contributes distinctly to the overall performance. Greenhouse-aware HSV augmentation delivers the largest individual improvement (+2.02 pp mAP@50), confirming that domain-specific color perturbation calibrated to the tomato ripeness hue range is more effective than generic COCO augmentation for chrominance-dependent detection~\citep{Rizzo2023}. The 3-phase progressive unfreezing strategy achieves the highest mAP@50:95 (64.6\%), demonstrating that localization quality benefits most from gradual backbone adaptation --- a metric directly relevant to center-point accuracy, as tighter bounding boxes yield more precise geometric center estimates. The BN channel pruning validates that the YOLO26n architecture is over-parameterized for binary agricultural classification: 30\% of channels can be removed with only 0.01 pp mAP@50 loss, reducing effective parameters from 2.58M to approximately 1.8M.

Among all eight evaluated models, YOLO26-RipeLoc Lite achieves the highest precision at 95.2\%, exceeding YOLO12n (92.3\%) by 2.9 pp. This advantage carries direct operational significance: the 95.2\% precision implies approximately 4.8 wasted cycles per 100 detections, compared to 7.7 for YOLO12n, a 37\% reduction in unproductive manipulation time. The center-point localization evaluation confirms that the CPL module achieves sufficient accuracy for direct robotic grasping guidance. The Euclidean RMSE of 4.86 pixels (3.80~mm at 500~mm working distance) is comparable to or better than the localization accuracy reported in dedicated picking-point estimation pipelines: \citet{Rong2023} reported 4--6~mm picking-point error using a separate semantic segmentation and morphological processing pipeline, while \citet{Bai2023} achieved 5--8~mm accuracy for clustered tomato grasping targets using multi-stage image analysis. Critically, the proposed CPL module operates within the same single forward pass as detection and classification, adding negligible latency ($<$0.5~ms for Gaussian refinement arithmetic) compared to the 15--30~ms overhead of standalone picking-point estimation modules.

The parameter efficiency of YOLO26-RipeLoc Lite (2.38M parameters, $\sim$1.8M after pruning, $\sim$4.5~GFLOPs) results in a model footprint that is well-suited for edge deployment. While all experiments in this study were conducted on the NVIDIA RTX~4090, the model's computational profile is comparable to architectures that have been successfully deployed on embedded platforms in prior agricultural robotics work~\citep{Tang2020}. Empirical validation of inference latency on the target NVIDIA Jetson Orin platform with TensorRT optimization is planned as immediate future work prior to field deployment.

\begin{figure}[!ht]
\centering
\includegraphics[width=0.4\linewidth]{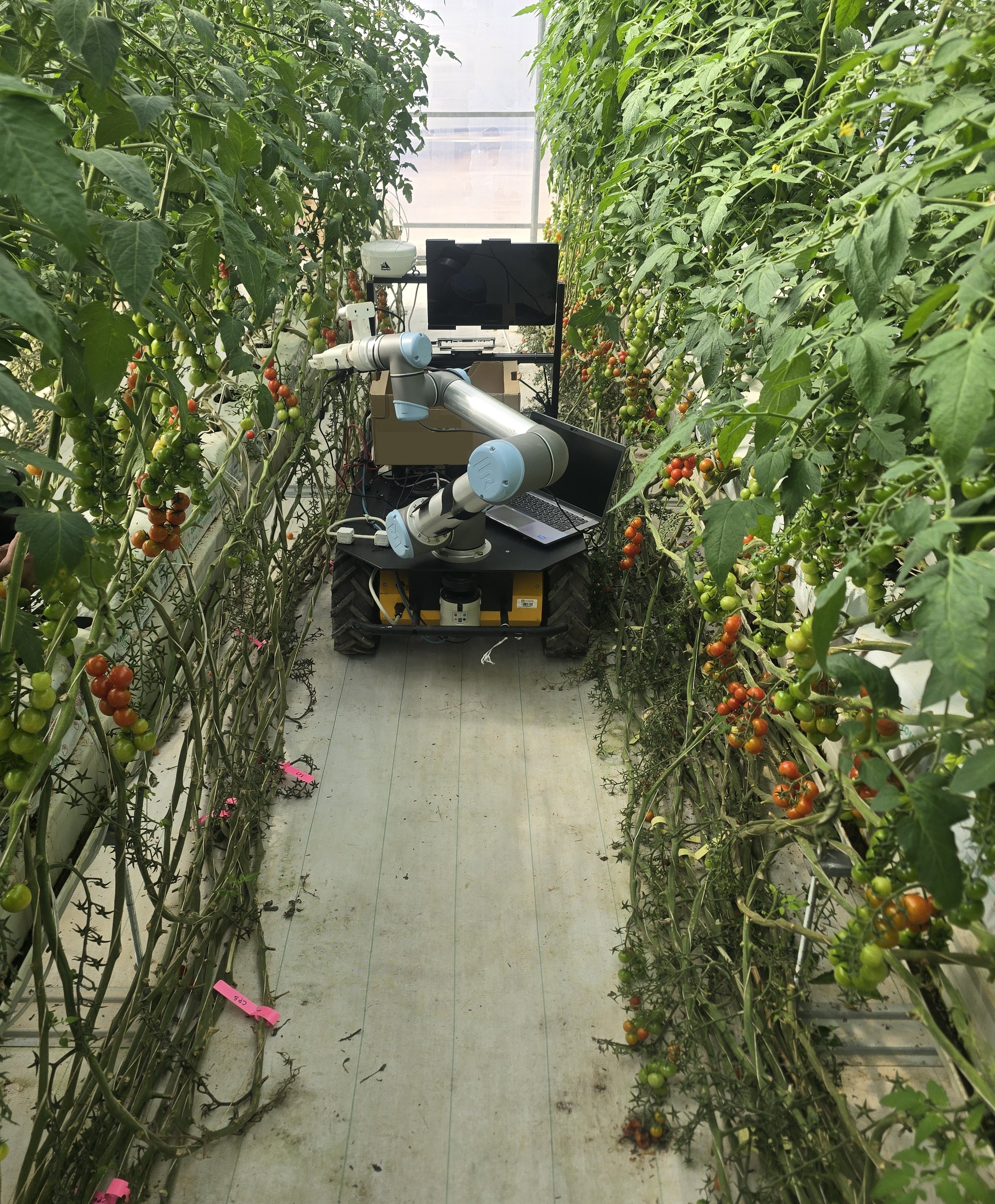}
\caption{Future scope: Hardware platform (Husky mobile robot + UR5 robotic arm with harvesting setup).}
\label{fig:fig11}
\end{figure}

Several limitations warrant acknowledgment. The dataset originates from a single greenhouse facility, and generalization to different cultivars and climate zones requires multi-site validation. The confusion matrix reveals that 76\% of background false positives are classified as unripe, indicating difficulty distinguishing green fruit from green vegetation; incorporating background images and hard negative mining would address this. The center-point localization relies on bounding box geometry, which introduces systematic offset for heavily occluded fruit; instance segmentation or keypoint regression would improve accuracy in such cases. The 9\% ripe-to-unripe confusion reflects residual breaker-stage ambiguity that could be addressed through a third class label or near-infrared imaging. Finally, deployment on the real robotic hardware requires integration with the RealSense D435 depth channel for 3D picking-point estimation, which is planned as immediate future work as depicted in Fig.~\ref{fig:fig11}.

\section{Conclusions}
\label{sec:conclusions}

This paper presented YOLO26-RipeLoc Lite, a lightweight single-stage detection architecture for simultaneous tomato ripeness detection and picking-point localization in greenhouse robotic harvesting. The proposed model introduces three task-specific modifications to the YOLO26 framework: a Lightweight Feature Pyramid Network (LFPN) with depthwise separable convolutions, a Ripeness-Aware Attention Module (RAAM) with learnable chrominance bias for ripe/unripe discrimination, and a Compact Detection Head (CDH) with shared convolutions and integrated center-point regression.

Experimental evaluation on a 1,500-image greenhouse tomato dataset (6,227 instances) demonstrated the following key findings:

\begin{enumerate}
    \item The proposed model achieves an overall mAP@50 of 92.9\% with per-class AP of 95.2\% (ripe) and 90.6\% (unripe), attaining the highest precision (95.2\%) among all eight evaluated YOLO variants while using the fewest parameters (2.38M).

    \item Greenhouse-aware HSV augmentation calibrated to the tomato ripeness hue range contributes the largest individual performance improvement (+2.02 pp mAP@50), confirming that domain-specific color augmentation is critical for chrominance-dependent agricultural detection tasks.

    \item The 3-phase progressive unfreezing training strategy achieves the best localization quality (mAP@50:95 of 64.6\%) by balancing pretrained feature preservation with domain adaptation.

    \item Post-training BatchNorm channel pruning at 30\% reduces effective parameters from 2.58M to approximately 1.8M with negligible accuracy loss (0.01 pp mAP@50).

    \item The integrated center-point localization module achieves a Euclidean RMSE of 4.86 pixels (3.80~mm at 500~mm working distance), with 79.4\% of predictions falling within the 5-pixel threshold, well within the $\pm$5~mm grasping tolerance of pneumatic end-effectors.

    \item The CPL module provides grasping coordinates for 126 ripe detections across 46 of 48 test images (95.8\% image-level coverage) with mean confidence of 0.856, enabling direct robotic end-effector guidance without a separate pose estimation pipeline.

    \item Comparative evaluation against YOLOv8, YOLO11, YOLO12, and YOLO26 baselines at matched scales establishes the first YOLO26-based benchmark for greenhouse tomato detection, demonstrating that the proposed architecture occupies a distinct position on the accuracy--efficiency Pareto frontier.
\end{enumerate}

Future work will focus on four directions: (1) real-time inference benchmarking on the NVIDIA Jetson Orin with TensorRT FP16/INT8 optimization, followed by integration with the Intel RealSense D435 depth channel for 3D center-point estimation and deployment on the physical Husky A200 + UR5 harvesting platform; (2) extension to multi-cultivar detection including cherry, grape, and beefsteak tomato varieties through transfer learning; (3) incorporation of instance segmentation for improved center-point accuracy under heavy occlusion; and (4) multi-site validation across geographically diverse greenhouse facilities to establish cross-domain generalization performance.


\section*{CRediT authorship contribution statement}

\textbf{Rajmeet Singh:} Conceptualization, Methodology, Formal analysis, Investigation, Writing -- original draft, Writing -- review \& editing.
\textbf{Manveen Kaur:} Data curation, Augmentation, Labeling.
\textbf{Shahpour Alirezaee:} Supervision, Resources, Writing -- review \& editing.
\textbf{Irfan Hussain:} Conceptualization, Supervision, Funding acquisition, Resources, Writing -- review \& editing.

\section*{Declaration of competing interest}

The authors declare that they have no known competing financial interests or personal relationships that could have appeared to influence the work reported in this paper.

\section*{Data availability}

Data will be made available on request.

\section*{Acknowledgements}

This work was supported in part by the Khalifa University Center for Autonomous Robotic Systems (KUCARS) Theme 4 under the Award RC1-2018-KUCARS, and in part by Silal Innovation Oasis through the projects under grants 8475000024.

\bibliography{ref}

\end{document}